\documentclass[10pt,twocolumn,letterpaper]{article}

\usepackage{cvpr}
\usepackage{times}
\usepackage{epsfig}
\usepackage{graphicx}
\usepackage{amsmath}
\usepackage{amssymb}
\usepackage{booktabs,multirow,array}
\usepackage{gensymb}
\usepackage{enumitem}
\setlist[itemize]{noitemsep, topsep=0pt}

\usepackage[pagebackref=true,breaklinks=true,letterpaper=true,colorlinks,bookmarks=false]{hyperref}

\cvprfinalcopy 

\def\cvprPaperID{****} 

\ifcvprfinal\fi
\begin{document}

\title{Toronto-3D: A Large-scale Mobile LiDAR Dataset for Semantic Segmentation of Urban Roadways}

\author{Weikai Tan\textsuperscript{1}, Nannan Qin\textsuperscript{1,2}, Lingfei Ma\textsuperscript{1}, Ying Li\textsuperscript{1},  Jing Du\textsuperscript{3}, Guorong Cai\textsuperscript{3}, Ke Yang\textsuperscript{4}, Jonathan Li\textsuperscript{1,4*}\\
\textsuperscript{1}Department of Geography and Environmental Management, \\
University of Waterloo, Waterloo, ON Canada N2L 3G1\\
\textsuperscript{2}Key Laboratory of Planetary Sciences, Purple Mountain Observatory, \\
Chinese Academy of Sciences, Nanjing, JS 210033, China\\
\textsuperscript{3}College of Computer Engineering, Jimei University, Xiamen, FJ 361021, China\\
\textsuperscript{4}Department of Systems Design Engineering, 
University of Waterloo, Waterloo, ON Canada N2L 3G1\\
{\tt\small \{weikai.tan, nannan.qin, l53ma, y2424li, ke.yang,  junli\textsuperscript{*}\}@uwaterloo.ca},\\
{\tt\small jingdu@jmu.edu.cn,  guorongcai.jmu@gmail.com}
}

\maketitle

\begin{abstract}
Semantic segmentation of large-scale outdoor point clouds is essential for urban scene understanding in various applications, especially autonomous driving and urban high-definition (HD) mapping. 
With rapid developments of mobile laser scanning (MLS) systems, massive point clouds are available for scene understanding, but publicly accessible large-scale labeled datasets, which are essential for developing learning-based methods, are still limited. 
This paper introduces Toronto-3D, a large-scale urban outdoor point cloud dataset acquired by a MLS system in Toronto, Canada for semantic segmentation. 
This dataset covers approximately 1 km of point clouds and consists of about 78.3 million points with 8 labeled object classes. 
Baseline experiments for semantic segmentation were conducted and the results confirmed the capability of this dataset to train deep learning models effectively. 
Toronto-3D is released \footnote{\url{https://github.com/WeikaiTan/Toronto-3D}} to encourage new research, and the labels will be improved and updated with feedback from the research community.
\end{abstract}

\section{Introduction}
Accurate and efficient scene perception of urban environments are crucial for various applications, including HD mapping, autonomous driving, 3D model reconstruction, and smart city \cite{campbell2010autonomous}. 
In the past decade, the largest portion of research in urban mapping is using 2D satellite and airborne imagery \cite{crommelinck2016review, ma2017review}, and autonomous driving researches also relied heavily on 2D images captured by digital cameras \cite{ros2015vision}. 
Compared with 2D images that are short of georeferenced 3D information, 3D point clouds collected by Light Detection and Ranging (LiDAR) sensors have become desirable for urban studies \cite{ma2019multi, remondino2011heritage}.
However, point clouds are unstructured, unordered and are usually in a large volume \cite{qi2017pointnet}. Deep learning algorithms have shown advantages tackling these challenges in point cloud processing in various tasks, including semantic segmentation \cite{boulch2017unstructured,qi2017pointnet++}, object detection \cite{chen20153d,zhou2018voxelnet}, classification \cite{luciano2018deep, qi2016volumetric}, and localization \cite{elbaz20173d,wang2018dels}. \\
Mobile platforms that integrate MLS sensors, location sensors (e.g. Global Navigation Satellite Systems (GNSS)), and 2D cameras (e.g. panoramic and digital cameras) are gaining popularity in urban mapping and autonomous driving due to the flexibility of data collection \cite{levinson2011towards, yang2015hierarchical}, but training effective deep learning models is not feasible without high-quality labels of the point clouds \cite{behley2019semantickitti}. 
The development of deep learning has always been driven by high-quality datasets and benchmarks \cite{torralba2011unbiased}. They allow researchers to focus on improving performance of algorithms without the hassle of collecting, cleaning and labeling large amount of data. They also ensure the performance of the algorithms are comparable with each other. \\
In this paper, Toronto-3D, a new large-scale urban outdoor point cloud dataset acquired by a MLS system is presented.
\begin{figure*}[ht]
\begin{minipage}{\textwidth}
  \centering
  \includegraphics[width=.98\textwidth]{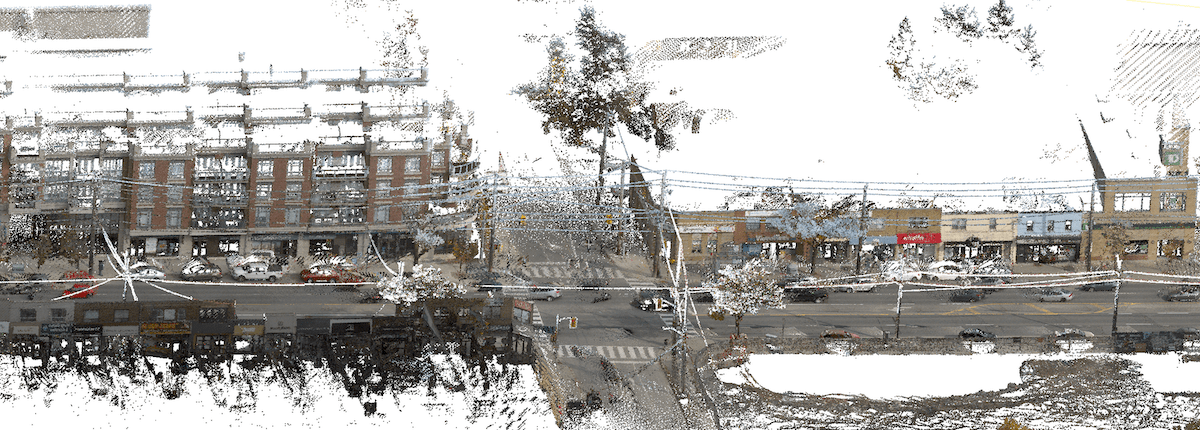}
\end{minipage}
\begin{minipage}{\textwidth}
  \centering
  \vspace{0.3cm}
  \includegraphics[width=.98\textwidth]{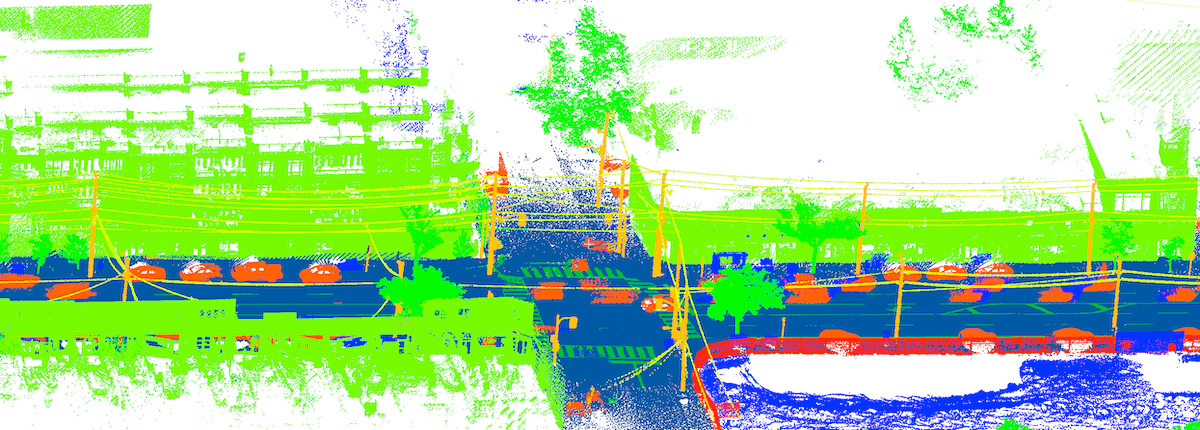}
\end{minipage}
\begin{minipage}{\textwidth}
  \centering
  \vspace{0.3cm}
  \includegraphics[width=.8\textwidth]{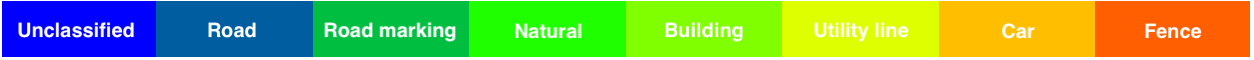}
\end{minipage}
\vspace{0.1cm}
\caption{Part of our dataset. 
    Top: dataset with natural color (RGB). 
    Bottom: class labels}
\label{fig:sample}
\end{figure*}
This dataset covers 1km of streets and consists of about 78.3 million points. A sample of the proposed dataset is shown in Fig. \ref{fig:sample}. The main contributions of this paper are to: 
\begin{itemize}
    \item present a large-scale point-wise labeled urban outdoor point cloud dataset for semantic segmentation, 
    \item investigate an integrated network for point cloud semantic segmentation, 
    \item provide an extensive comparison on the performance of state-of-the-art deep learning semantic segmentation methods on the proposed dataset.
\end{itemize}
%
\section{Available point cloud datasets for 3D Semantic Segmentation}
With the advancement of LiDAR and RGB-D sensors, and the development of autonomous driving and 3D vision,  point cloud data has become more and more accessible. 
However, such datasets usually have a very large volume of data and contain lots of noise, making it difficult and time-consuming to produce high-quality manual labels. 
Popular accessible outdoor point cloud datasets for semantic segmentation are as follows:
\begin{table*}[ht]
\caption{Recent urban outdoor point cloud datasets for semantic segmentation}
\label{tab:datasets}
\vspace{0.2cm} 
\begin{tabular*}{\textwidth}{cccccccc}
\toprule \midrule
Dataset & Year & Primary fields & Length & \# points & \begin{tabular}[c]{@{}c@{}}\# classes \\ labeled\end{tabular} & \begin{tabular}[c]{@{}c@{}}\# classes \\ evaluated\end{tabular} & \begin{tabular}[c]{@{}c@{}}LiDAR \\ Sensor\end{tabular} \\ \midrule
Oakland \cite{munoz2009contextual} & 2009 & x, y, z, label & 1510 m & 1.6 M & 44 & 5 & SICK LMS \\
iQmulus \cite{vallet2015terramobilita} & 2015 & \begin{tabular}[c]{@{}c@{}}x, y, z, intensity, \\ GPS time, scan origin, \\ \# echoes, object ID, label\end{tabular} & 200 m & 12 M & 22 & 8 & \begin{tabular}[c]{@{}c@{}}Riegl \\ LMS-Q120i\end{tabular} \\
Semantic3D \cite{hackel2017isprs} & 2017 & \begin{tabular}[c]{@{}c@{}}x, y, z, R, G, B, \\ intensity, label\end{tabular} & - & 4 B & 8 & 8 & \begin{tabular}[c]{@{}c@{}}Terrestrial \\ Laser Scanner\end{tabular} \\
Paris-Lille-3D \cite{roynard2017parisIJRR} & 2018 & x, y, z, intensity, label & 1940 m & 143.1 M & 50 & 9 & \begin{tabular}[c]{@{}c@{}}Velodyne \\ HDL-32E\end{tabular} \\
SemanticKITTI \cite{behley2019semantickitti} & 2019 & x, y, z, intensity, label & 39.2 km & 4.5 B & 28 & 25 & \begin{tabular}[c]{@{}c@{}}Velodyne \\ HDL-64E\end{tabular} \\ \midrule
\textbf{Toronto-3D (Ours)} & 2020 & \begin{tabular}[c]{@{}c@{}}x, y, z, R, G, B, \\ intensity, GPS time, \\ scan angle rank, label\end{tabular} & 1000 m & 78.3 M & 8 & 8 & \begin{tabular}[c]{@{}c@{}}Teledyne\\ Optech \\ Maverick\end{tabular}\\ \bottomrule
\end{tabular*}
\end{table*} 
\\
\textit{\textbf{Oakland 3-D}} \cite{munoz2009contextual} is one of the earliest outdoor point cloud datasets acquired by a MLS system mounted with a side-looking SICK LMS sensor. The sensor is a mono-fiber LiDAR, and the point density is relatively low. This dataset contains about 1.6 million points and was labeled into 44 classes. However, only 5 classes: vegetation, wire, pole, ground, and facade, were evaluated in literature. This dataset is relatively small so that it is more suitable for developing and testing lightweight networks.\\
\textit{\textbf{iQmulus}}  \cite{vallet2015terramobilita} dataset comes from the IQmulus \& TerraMobilita Contest acquired by a system called Stereopolis II \cite{paparoditis2012stereopolis} in Paris. A monofiber Riegl LMS-Q120i LiDAR was used to collect the point clouds. The full dataset has over 300 million points labeled into 22 classes, but only a small part of the dataset of 12 million points in a 200 m range with 8 valid classes was publicly available for the contest dataset. This dataset suffers unsatisfactory quality of classification due to occlusion from the monofiber LiDAR sensor and the annotation process \cite{roynard2017parisIJRR}. \\
\textit{\textbf{Semantic3D}} \cite{hackel2017isprs} is collected by terrestrial laser scanners, and it has much higher point density and accuracy compared with the other datasets. 8 class labels were included in this dataset. However, only very limited viewpoints are feasible for static laser scanners, and similar datasets are not easily acquired in practice.\\
\textit{\textbf{Paris-Lille-3D}} \cite{roynard2017parisIJRR} is one of the most popular outdoor point cloud datasets in recent years. The dataset was collected with a MLS system using Velodyne HDL-32E LiDAR, with a point density and measurement accuracy closer to point cloud data acquired by autonomous driving vehicles. The dataset covers close to 2km with over 140 million points, and very detailed labels of 50 classes were provided. For benchmarks, the dataset uses 9 classes for the purpose of semantic segmentation.\\
\textit{\textbf{SemanticKITTI}} \cite{behley2019semantickitti} is one of the most recent and largest publicly available datasets serving the purpose of semantic segmentation. The dataset was further annotated on the widely used KITTI dataset \cite{geiger2012we}. This dataset contains about 4.5 billion points covering close to 40 km, and it is labeled by each sequential scan with 25 classes for the evaluation of semantic segmentation. This dataset is more focused on algorithms towards autonomous driving. \\
Development and validation of deep learning algorithms demand more datasets with various object labels. Toronto-3D is introduced in this paper to provide an additional high-quality point cloud dataset for 3D semantic segmentation with new labels. Table \ref{tab:datasets} shows a comparison of comprehensive indicators with the above-mentioned datasets.
%
\section{New dataset: Toronto-3D}
\subsection{Data acquisition}
The point clouds in this dataset were acquired with a vehicle-mounted MLS system: Teledyne Optech Maverick\footnote{\url{http://www.teledyneoptech.com/en/products/mobile-survey/maverick/}}. The system consists of a 32-line LiDAR sensor, a Ladybug 5 panoramic camera, a GNSS system, and a Simultaneous Localization and Mapping (SLAM) system. The LiDAR sensor can capture point clouds at up to 700,000 points per second at a vertical field of view covering from -10\degree to +30\degree, with an accuracy of better than 3 cm. The collected point clouds were further processed with LMS Pro\footnote{\url{https://www.teledyneoptech.com/en/products/software/lms-pro/}} software. Natural color (RGB) was assigned to each point with reference to the imaging camera.
%
%
\subsection{Description of the dataset}
\begin{figure*}[t]
\begin{minipage}{\textwidth}
  \centering
  \includegraphics[width=.98\textwidth]{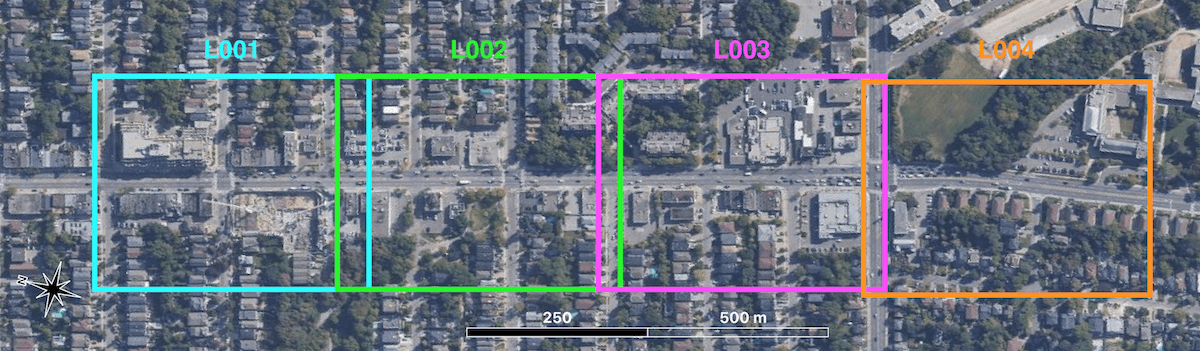}
\end{minipage}
\vspace{0.2cm}
\begin{minipage}{\textwidth}
  \flushright
  \includegraphics[width=.95\textwidth]{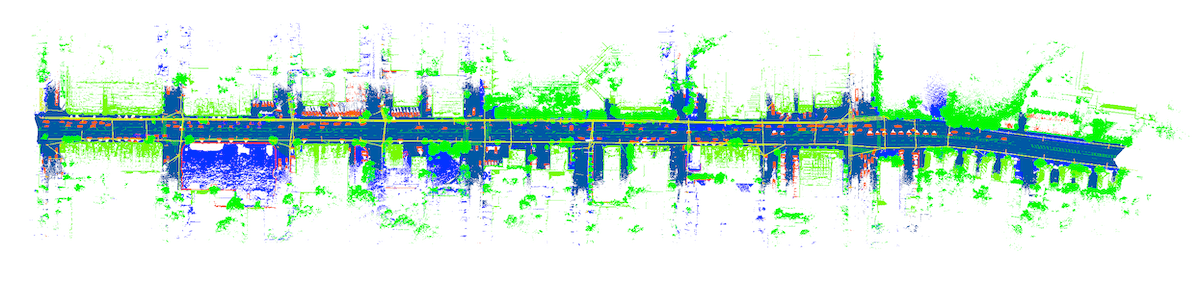}
\end{minipage}
\vspace{0.1cm}
\begin{minipage}{\textwidth}
  \centering
  \includegraphics[width=.8\textwidth]{Figures/colorbar.png}
\end{minipage}
\vspace{0.1cm}
\caption{Overview of the dataset. 
    Top: Approximate boundary of each section in our dataset (Satellite imagery from Google Maps). 
    Bottom: Overview of labels (each class in different colors).}
\label{fig:overview}
\end{figure*}
This dataset was collected on Avenue Road in Toronto, Canada, covering approximately 1 km of road segment with approximately 78.3 million points. This dataset is divided into four sections, and each section covers a range of about 250 m. An overview of the approximate boundary of each section is illustrated in Fig. \ref{fig:overview}. 
This dataset is collected using a 32-line LiDAR sensor, and the point clouds have high density of about 1000 points/m\textsuperscript{2} on the ground on average. The dataset covers the full range of the MLS sensor of approximately 100 m away from the road centerline without trimming. Limited post-processing was done to resemble real-world point cloud collection scenarios without trimming far points and resampling.\\
Each of the four sections of the dataset was saved separately in \textit{.ply} files. The point clouds were classified and point-wise labels were assigned manually using CloudCompare\footnote{\url{https://www.cloudcompare.org}} software. Each point cloud file has the following 10 attributes:
\begin{itemize}
    \item \textit{x, y, z}: Position of each point recorded in meters, in NAD83 / UTM Zone 17N
    \item \textit{R, G, B}: Natural color reflectance of red, green, blue of each point, recorded in integer [0, 255]
    \item \textit{Intensity}: LiDAR intensity of each point, normalized to integer [0, 255]
    \item \textit{GPS time}: GPS time of when each point was collected, recorded in float format
    \item \textit{Scan Angle Rank}: Scan angle of each point in degree, recorded in integer [-13, 31]
    \item \textit{Label}: Object class label of each point, recorded in integer [0, 8]
\end{itemize}
A sample of the Toronto-3D dataset is shown Fig. \ref{fig:sample}. Similar to previous datasets, the object class labels were defined as follows:
\begin{itemize}
    \item \textit{Road (label 1)}: Paved road surfaces, including sidewalks, curbs, parking lots
    \item \textit{Road marking (label 2)}: Pavement markings including driving lines, arrows, pedestrian crossings
    \item \textit{Natural (label 3)}: Trees, shrubs, not including grass and bare soil
    \item \textit{Building (label 4)}: Any parts of low and multi-story buildings, store fronts
    \item \textit{Utility line (label 5)}: Power lines, telecommunication lines over the streets
    \item \textit{Pole (label 6)}: Utility poles, traffic signs, lamp posts
    \item \textit{Car (label 7)}: Moving cars and parked cars on road sides and parking lots
    \item \textit{Fence (label 8)}: Vertical barriers, including wooden fences, walls of construction sites
    \item \textit{unclassified (label 0)}
\end{itemize}
A summary of number of points and distribution of labels in each section is shown in Table \ref{tab:npts}.
\begin{table*}[t]
\caption{Number of labeled points for each class (thousand)}
\label{tab:npts}
\vspace{0.2cm}
\begin{tabular*}{\textwidth}{ccccccccccc}
\toprule \midrule
Section & Road & Road marking & Natural & Building & Utility line & Pole & Car & Fence & unclassified & Total \\ \midrule
L001 & 11,178 & 433 & 1,408 & 6,037 & 210 & 263 & 1,564 & 83 & 391 & 21,567 \\
L002 & 6,353 & 301 & 1,942 & 866 & 84 & 155 & 199 & 24 & 360 & 10,284 \\
L003 & 20,587 & 786 & 1,908 & 11,672 & 332 & 408 & 1,969 & 300 & 1,760 & 39,722 \\
L004 & 3,738 & 281 & 1,310 & 525 & 37 & 71 & 200 & 4 & 582 & 6,748 \\ \midrule
Total & 41,856 & 1,801 & 6,568 & 19,100 & 663 & 897 & 3,932 & 411 & 3,093 & 78,321\\
\bottomrule
\end{tabular*}
\end{table*}
\subsection{Challenges of Toronto-3D}
\begin{figure}[t]
\begin{minipage}{\linewidth}
  \centering
  \includegraphics[width=.85\linewidth]{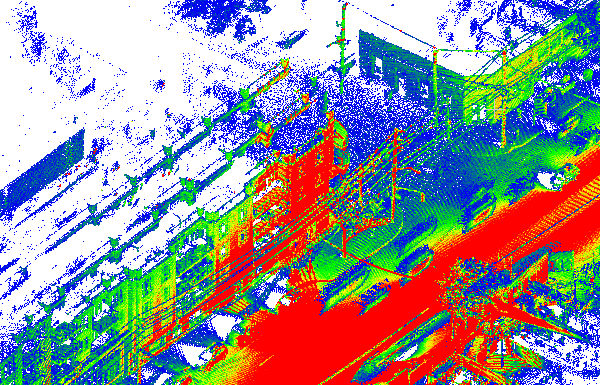}
  \vspace{0.2cm}
\end{minipage}
\begin{minipage}{\linewidth}
  \centering
  \includegraphics[width=.85\linewidth]{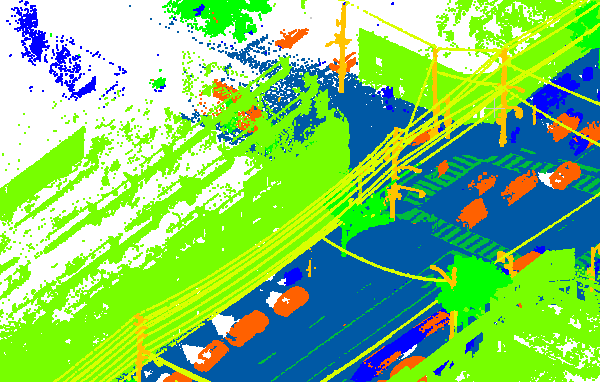}
\end{minipage}
\vspace{0.1cm}
\caption{Variations of point density. 
    Top: Point density high to low illustrated in color red to blue. 
    Bottom: Labels of point cloud (each class in different colors).}
\label{fig:density}
\end{figure}

The Toronto-3D dataset is comparable to Paris-Lille-3D in several aspects. 
They are both urban outdoor large-scale scenes collected by a vehicle-mounted MLS system with a 32-line LiDAR. 
Toronto-3D covers approximately half the distance of Paris-Lille-3D and includes half the number of points. 
They are both labeled with a similar number of classes for the purpose of semantic segmentation. 
Different from Paris-Lille-3D, the Toronto-3D dataset has the following characteristics that bring more challenges to effective point cloud semantic segmentation algorithms.\\
\textit{\textbf{Full coverage of LiDAR measurement range.}} The MLS system that acquired this dataset has a valid measurement distance of approximately 100 m. Different from Paris-Lille-3D where only points within approximately 20 m away from the road centerline are available, Toronto-3D keeps all collected points within about 100 m without trimming. The full coverage of measurement range of Toronto-3D resembles point cloud data collection in real-world scenarios, and it brings challenges of variations of point density at different distances,  inclusion of more noise, and inclusion of more objects further away from the sensor. \\
\textit{\textbf{Variation of point density.}} Unlike the relatively small variation of point density in Paris-Lille-3D, the Toronto-3D dataset has a larger variation of point density of objects caused majorly by two reasons: inclusion of all points within full LiDAR measurement range, and repeated scans during point cloud collection. The variations of point density are illustrated in Fig. \ref{fig:density}. As illustrated in the scene, the cars (colored in orange) on the streets have much higher point density compared to the parked cars at the upper-middle in the image. The cars with lower density are approximately 30-40 m away from the road centerline, which means such scenarios would not be included in Paris-Lille-3D. In addition, at the center area in the scene, point density is significantly higher (over 10 times higher) compared to other parts of the scene, and this is caused by repeated scans when the vehicle stopped at the intersection during data collection. The repeated scans resulted in variations of point density on the same building at the same distance to the sensor at different locations, and no resampling process was performed. The large variation of point density would be challenging to test the robustness of algorithms to capture features effectively.\\
\textit{\textbf{New challenging classes.}} There are two class labels not commonly seen in other popular datasets listed in Table \ref{tab:datasets}, i.e., road marking and utility line. 
Road markings include various pavement markings on the road surface, including pedestrian crossings and lane arrows, with various sizes and shapes, and they are difficult to distinguish from road surfaces. 
Though wires, defined similar to utility lines this dataset, were included in Oakland 3-D dataset \cite{munoz2009contextual}, sample size was limited due to the small number of points in the dataset. 
The utility lines are thin linear objects that are challenging to identify, especially in areas where they overlap with poles, trees and are close to buildings. 
In addition, the fence class that covers various wall-like vertical structures is also challenging to identify.
%
\section{Methods}
\subsection{Recent studies}
Semantic segmentation of point clouds is to make predictions on each point to assign a semantic label. With the recent development of 3D deep learning, semantic segmentation tasks can be achieved by end-to-end deep neural networks. Existing 3D deep learning models on point clouds can be roughly generalized into three categories: view-based models, voxel-based models and point-based models. \\
The view-based models such as MVCNN \cite{su2015multi} project 3D point clouds into multiple views as 2D images, 
but they do not fully use the rich 3D information. 
Voxel-based models such as VoxNet \cite{maturana2015voxnet} and 3D-CNN \cite{huang2016point} structure unordered point clouds into voxel grids, so that known structures and methods of 2D images can be extended to 3D space. However, the nature of point clouds that they are sparse and have varying densities make voxelization inefficient. \\
Point-based methods directly process unordered and unstructured point clouds to capture 3D spatial features. 
Starting from PointNet \cite{qi2017pointnet} which learns point-wise spatial feature with multi-layer perceptron (MLP) layers, point-based methods have been greatly developed, followed by PointNet++ \cite{qi2017pointnet++}, PointCNN \cite{li2018pointcnn}. Graph models were also applied to extract spatial features in point-based models, and such methods include ECC \cite{simonovsky2017dynamic} and DGCNN \cite{wang2019dynamic}. 
\subsection{Baseline approaches for semantic segmentation}
Six state-of-the-art point-based deep learning models for semantic segmentation were tested on the proposed dataset as baseline approaches:\\
\textit{\textbf{PointNet++}} \cite{qi2017pointnet++} applies PointNet \cite{qi2017pointnet}, which is the pioneer method using MLPs to process point cloud directly, to local neighborhoods of each point to capture local features, and a hierarchical approach is taken to capture both local and global features. \\
\textit{\textbf{DGCNN}} \cite{wang2019dynamic} constructs graphs to extract local geometric features from local neighborhoods, and applies EdgeConv as a convolution-like operation. EdgeConv is isotropic about input features with convolutional operations on graph nodes and their edges.\\
\textit{\textbf{KPFCNN}} \cite{thomas2019KPConv} introduces a convolutional operator called KPConv to capture local features with weights defined by a set of kernel points. KPConv is robust to varying point densities and is computationally efficient. KPFCNN is currently ranked first in Paris-Lille-3D benchmark.\\
\textit{\textbf{MS-PCNN}} \cite{ma2019multi} is an end-to-end point cloud segmentation network combining point convolutions with edge information. It applies revised PointConv \cite{wu2019pointconv} operations with edge features from revised EdgeConv \cite{wang2019dynamic} operations. \\
\textit{\textbf{TGNet}} \cite{li2019tgnet} introduces a novel graph convolution function called TGConv defined as products of point features from local neighborhoods. The features are extracted with Gaussian weighted Taylor kernel functions. It is an end-to-end semantic segmentation network with hierarchical TGConv followed by a conditional random field (CRF) layer.\\
\textit{\textbf{MS-TGNet}} is proposed in this study with a revised structure of TGNet.
Considering the full range of approximately 100 m from the road centerline was preserved in this dataset, there is a large difference in point density. 
Multi-scale grouping (MSG) proposed in PointNet++ \cite{qi2017pointnet++} was designed to capture features more effectively in point clouds with large variations in point density.
A MSG layer was implemented in the second layer of the original TGNet architecture to capture local geometric features at three different radii (0.2 m, 0.4 m and 0.8 m) through testing. 
\subsection{Evaluation metrics}
For the evaluation of semantic segmentation results, intersection over union (\(IoU\)) of each class, overall accuracy (\(OA\)) and mean IoU (\(mIoU\)) are used.
\begin{equation}
    \label{eq1}
    IoU_n=\frac{TP_n}{TP_n+FP_n+FN_n}
\end{equation}
\begin{equation}
    \label{eq2}
    OA=\frac{\sum TP_n}{Total\ number\ of\ points}
\end{equation}
\begin{equation}
    \label{eq3}
    mIoU=\frac{\sum IoU_n}{N}
\end{equation}
where \(N\) is the total number of labels, \(n\) is the \(n\)th label in \(N\), \(TP\), \(FP\) and \(FN\) represent number of points of true positives, false positives and false negatives of the predictions respectively. \(OA\) and \(mIoU\) evaluate the overall quality of semantic segmentation, and \(IoU\) of each class measures the performance on each class.
\subsection{Parameters and configurations}
L002 was selected as the testing set among the four sections due to its smaller size and balanced number of points of each label, while the other three sections were used for training and validation. 
For fair comparison, only coordinates (\(x,y,z\)) of point clouds were used.\\
The parameter settings of PointNet++ and DGCNN were directly used from the networks for indoor scenes, which may limit the performance of these two algorithms to some extent. The parameter settings of KPFCNN, MS-PCNN and TGNet were used from the networks tested on Paris-Lille-3D dataset in literature.
The network structures and parameter settings of these algorithms may not be directly comparable, and parameter tuning does not guarantee the fairness of comparison.
In this study, the results are for baseline illustration purpose only, and better results could be potentially achieved with further tuning.
The models were trained and tested on a NVIDIA RTX 2080Ti with 11G of RAM, and batch sizes were adjusted accordingly. 
\section{Results and discussions}
\subsection{Performance of baseline approaches}
\begin{figure*}[ht]
\begin{minipage}{\textwidth}
  \centering
  \includegraphics[width=.8\textwidth]{Figures/colorbar.png}
  \vspace{0.2cm}
\end{minipage}
\begin{minipage}[t]{.33\textwidth}
  \centering
  \includegraphics[width=\textwidth]{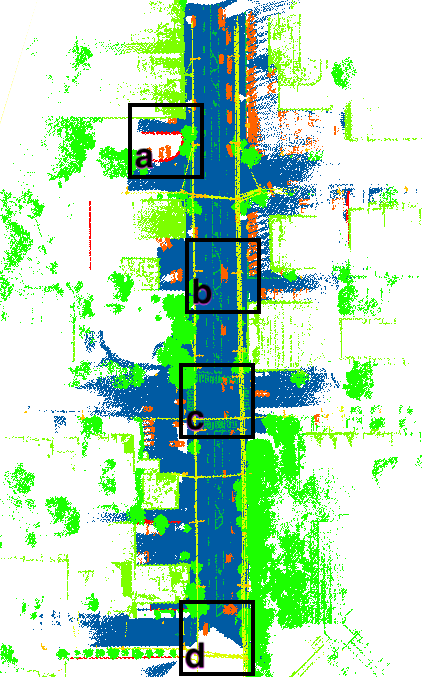}
  \centerline{(a) Ground truth}
\end{minipage}
\hfill
\begin{minipage}[t]{.33\textwidth}
  \centering
  \includegraphics[width=\textwidth]{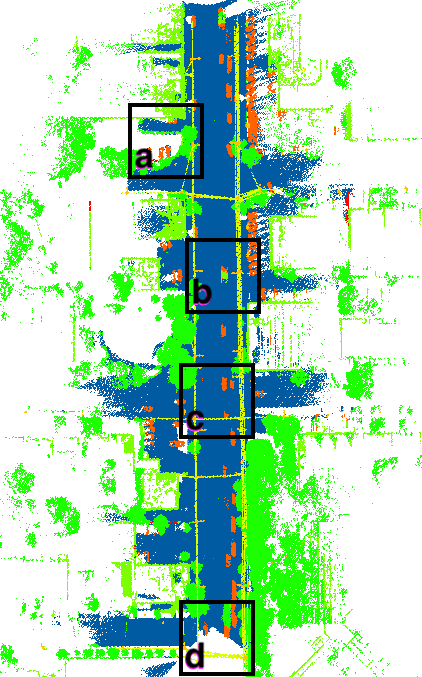}
  \centerline{(b) Result of KPFCNN}
\end{minipage}
\hfill
\begin{minipage}[t]{.33\textwidth}
  \centering
  \includegraphics[width=\textwidth]{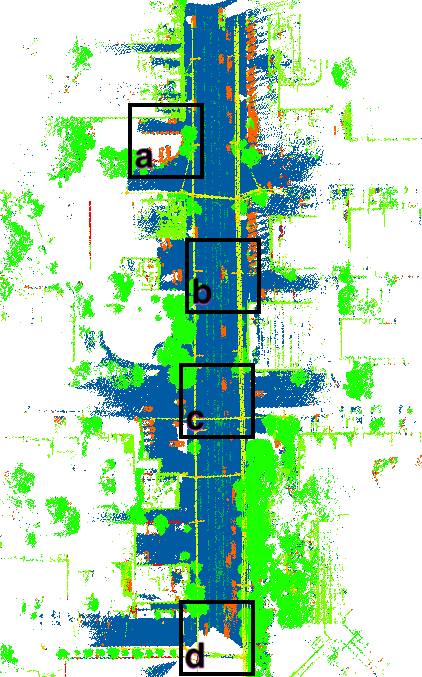}
  \centerline{(c) Result of MS-TGNet}
\end{minipage}
\vspace{0.2cm}
\caption{Visual comparison of results of semantic segmentation}
\label{fig:results}
\end{figure*}

\begin{table*}[ht]
\caption{Semantic segmentation results of different methods (\%)}
\label{tab:baseline}
\vspace{0.2cm}
\begin{tabular*}{\textwidth}{ccccccccccc}
\toprule \midrule
Methods & \(OA\) & \(mIoU\) & Road & Rd mrk. & Natural & Building & Util. line & Pole & Car & Fence \\ \midrule
PointNet++ \cite{qi2017pointnet++} & 91.21 & 56.55 & \textbf{91.44} & 7.59 & 89.80 & 74.00 & 68.60 & 59.53 & 53.97 & 7.54 \\
PointNet++ (MSG) \cite{qi2017pointnet++} & 90.58 & 53.12 & 90.67 & 0.00 & 86.68 & 75.78 & 56.20 & 60.89 & 44.51 & 10.19 \\
DGCNN \cite{wang2019dynamic} & 89.00 & 49.60 & 90.63 & 0.44 & 81.25 & 63.95 & 47.05 & 56.86 & 49.26 & 7.32 \\
KPFCNN \cite{thomas2019KPConv} & \textbf{91.71} & 60.30 & 90.20 & 0.00 & 86.79 & \textbf{86.83} & \textbf{81.08} & \textbf{73.06} & 42.85 & \textbf{21.57} \\
MS-PCNN \cite{ma2019multi} & 91.53 & 58.01 & 91.22 & 3.50 & 90.48 & 77.30 & 62.30 & 68.54 & 53.63 & 17.12 \\
TGNet \cite{li2019tgnet} & 91.64 & 58.34 & 91.39 & 10.62 & 91.02 & 76.93 & 68.27 & 66.25 & \textbf{54.10} & 8.16 \\ \midrule
\textbf{MS-TGNet (Ours)} & 91.69 & \textbf{60.96} & 90.89 & \textbf{18.78} & \textbf{92.18} & 80.62 & 69.36 & 71.22 & 51.05 & 13.59 \\ \bottomrule
\end{tabular*}
\end{table*}
The results for semantic segmentation baseline approaches using Toronto-3D are shown in Table \ref{tab:baseline}. \\
PointNet++ \cite{qi2017pointnet++} achieved highest \(IoU\) in road class. However, the PointNet++ model with MSG modules did not perform as well as the base PointNet++ architecture with the published parameter settings of indoor scenes. DGCNN \cite{wang2019dynamic} performed the worst in terms of both \(OA\) and \(mIoU\) in our dataset. Since DGCNN uses KNN for construction of graphs to capture local features, it may not perform well in this dataset with varying point density.\\
KPFCNN \cite{thomas2019KPConv} is on the top spot of Paris-Lille-3D benchmark at the moment, and it achieved the highest \(OA\) and second highest \(mIoU\) among the tested baseline algorithms. 
KPFCNN achieved the highest \(IoU\) in building, utility line, pole and fence segmentation.
MS-PCNN \cite{ma2019multi} and TGNet \cite{li2019tgnet} both achieved \(mIoU\) of over 58\% following the performance of KPFCNN.\\
The proposed MS-TGNet achieved comparable results with KPFCNN, achieving  highest \(mIoU\) of 60.96\% and second highest \(OA\) of 91.69\% among the baseline approaches. It has highest \(IoU\)s in road marking and natural classes.
A visual comparison of semantic segmentation results of KPFCNN and MS-TGNet is shown in Fig. \ref{fig:results}. 
%
\newcolumntype{C}{>{\centering\arraybackslash}m{.196\textwidth}}
\begin{table*}[ht]
\caption{Detailed views at semantic segmentation errors}
\label{tab:zoomed}
\vspace{0.2cm}
\begin{tabular}{c*4{C}@{}}
\toprule \midrule
 & a & b & c & d \\ \midrule
KPFCNN & \includegraphics[width=.15\textwidth]{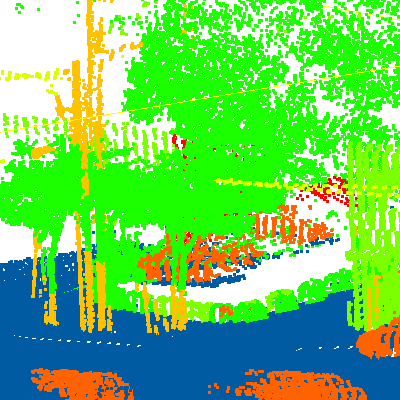} & 
\includegraphics[width=.15\textwidth]{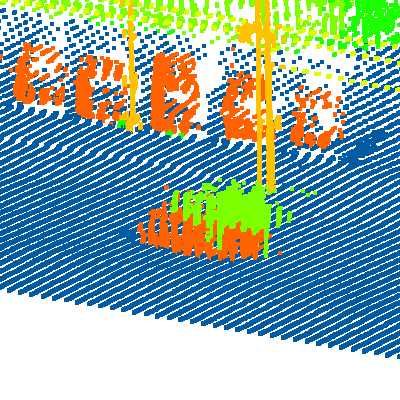} & 
\includegraphics[width=.15\textwidth]{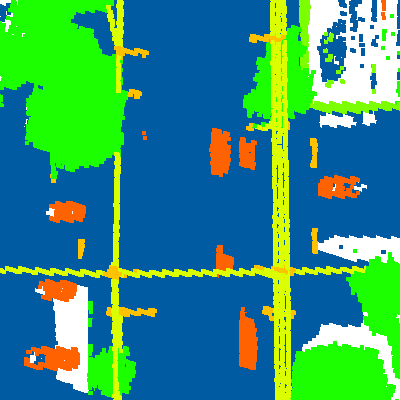} & 
\includegraphics[width=.15\textwidth]{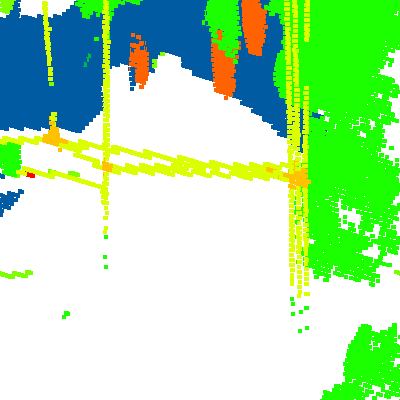} \\ 
MS-TGNet & \includegraphics[width=.15\textwidth]{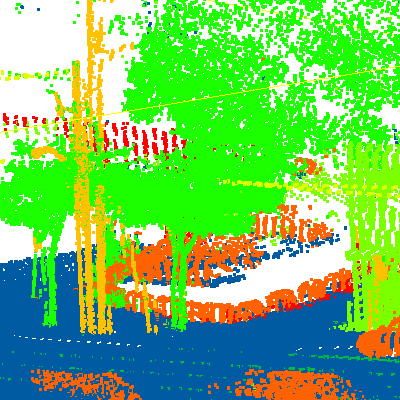} & 
\includegraphics[width=.15\textwidth]{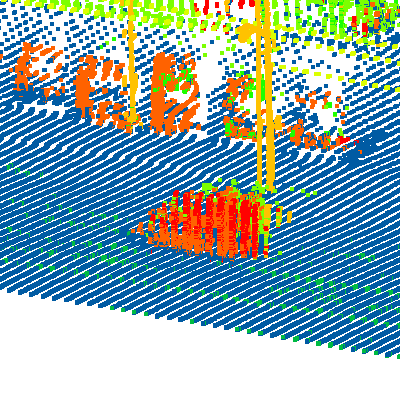} & 
\includegraphics[width=.15\textwidth]{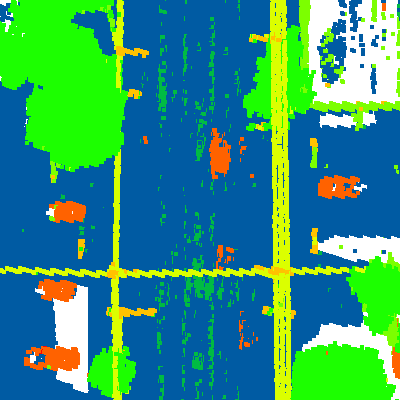} & 
\includegraphics[width=.15\textwidth]{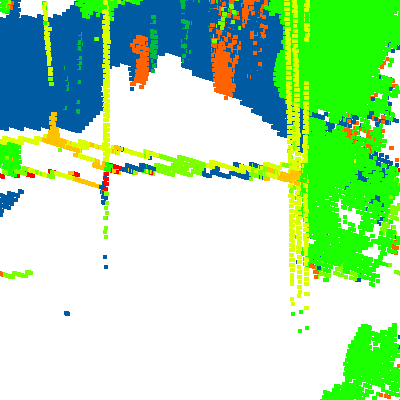} \\ 
Ground truth & \includegraphics[width=.15\textwidth]{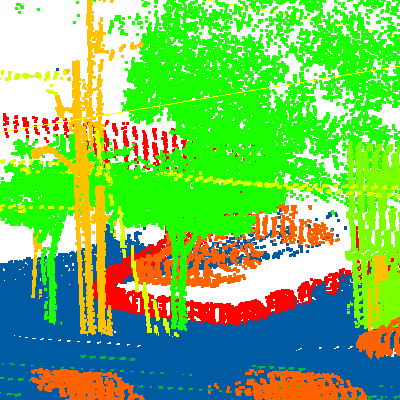} & 
\includegraphics[width=.15\textwidth]{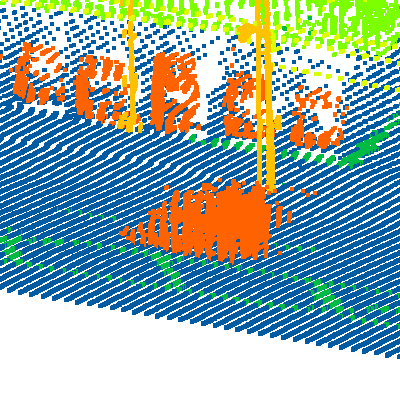} & 
\includegraphics[width=.15\textwidth]{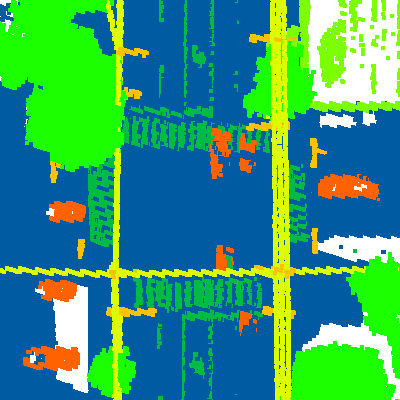} & 
\includegraphics[width=.15\textwidth]{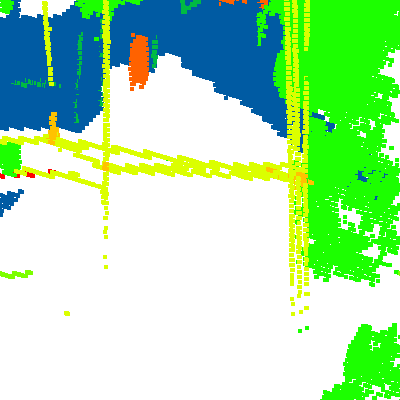} \\ 
\bottomrule 
\end{tabular}
\end{table*} 
\subsection{Areas for improvements}
From the results of the baseline approaches, all algorithms performed poorly in road marking and fence with \(IoU\)s lower than 22\%, and the accuracy of utility line, pole and car can be improved. 
Four examples of errors in the results from the best performing two algorithms, KPFCNN and MS-TGNet, are further illustrated in Table \ref{tab:zoomed}.\\
In \textit{Box a} where some concrete blocks are placed as a barrier, KPFCNN classified them as buildings while MS-TGNet classified them as cars. 
The concrete blocks may have similar structures to cars and buildings but they are smaller in size. 
In the same scene, part of trunks of the trees were misclassified as poles in KPFCNN but they are correctly grouped into natural class in MS-TGNet possibly due to the edge features. 
It explained the higher performance of KPFCNN on pole and higher performance of MS-TGNet on natural class to some extent. \\
\textit{Box b} shows a truck partially misclassified by both algorithms. 
They both correctly classified the lower part of the truck, but the upper part with box shape was classified as building by KPFCNN and fence by MS-TGNet. 
The truck is a moving object and only one side was completely scanned, adding the vertical structure of the truck, so that these conditions possibly resulted in the confusion.\\
\textit{Box c} shows a scenario with road markings. KPFCNN was not able to identify road markings with a \(IoU\) close to 0. MS-TGNet captured part of the road markings but at a low \(IoU\) of 18.78\%, and missed the pedestrian crossings. 
The road markings on pavements are difficult to distinguish with point coordinates only, and additional color and intensity information would make road markings easier to identify.\\
\textit{Box d} illustrates a scene with utility lines at the boundary of the test point cloud. KPFCNN distinguished the utility lines correctly, while MS-TGNet classified the horizontal line as a mixture of several classes including road and road markings.
The horizontal utility line in \textit{Box d} is at the boundary of the point cloud without road surface underneath to provide contextual information.\\
The performance of algorithms with the backbone of PointNet++, including the proposed MS-TGNet, was probably limited by the number of points the structure is able to process with the limitation of RAM. 
KPFCNN uses a much larger number of points and outperformed MS-TGNet in most categories but showed some weakness in natural and road marking classification. 
Selection of KPConv kernels and network settings could be improved to capture road marking.
New algorithms that can process a large number of points, such as RandLA-Net \cite{hu2019randla}, would have advantages in outdoor scenes, and they will be tested on Toronto-3D.
\section{Conclusions}
This paper presents Toronto-3D, a new large-scale urban outdoor point cloud dataset collected by a MLS system. 
The dataset covers approximately 1km of road with over 78 million points in Toronto, Canada. 
All points were preserved in the range of data collection to resemble real-world application scenarios.
This dataset was manually labeled into 8 categories, including road, road marking, natural, building, utility line, pole, car and fence. 
Five state-of-the-art end-to-end point cloud semantic segmentation algorithms and a proposed network named MS-TGNet were tested as baselines for this dataset. \\
The proposed MS-TGNet is able to produce comparative performance with state-of-the-art methods, achieving the highest \(mIoU\) of 60.96\% and a competitive \(OA\) of 91.69\% in the new dataset. 
The Toronto-3D dataset provides new class labels including road markings, utility lines and fences, and all tested semantic segmentation methods need to improve on road markings and fences.\\
The intention of presenting this new point cloud dataset is to encourage developing creative deep learning models. The labels of this new dataset will be improved and updated with feedback from the research community.
\section*{Acknowledgements}
Teledyne Optech is acknowledged for providing mobile LiDAR point cloud data. Thanks Jimei University for point cloud labeling.
{\small
\bibliographystyle{ieee_fullname}
\bibliography{2020_CVPR_EV}
}
\end{document}